\pdfoutput=1

\documentclass{article}
\usepackage{ijcai24}

\usepackage{times}
\usepackage{soul}
\usepackage{url}
\usepackage[hidelinks]{hyperref}
\usepackage[utf8]{inputenc}
\usepackage[small]{caption}
\usepackage{svg}
\usepackage{graphicx}
\usepackage{amsmath}
\usepackage{amsthm}
\usepackage{booktabs}
\usepackage{multirow}
\usepackage{algorithm}
\usepackage{algpseudocode}
\usepackage[switch]{lineno}
\usepackage[most]{tcolorbox}
\newtcolorbox{prompt_box}{colback=orange!5!white,colframe=orange!75!black}
\usepackage{textcomp}
\usepackage{xcolor}
\usepackage{soul}
\usepackage{enumitem}
\usepackage{amssymb}
\usepackage{arydshln}

\title{ECR-Chain: Advancing Generative Language Models to \\Better Emotion-Cause Reasoners through Reasoning Chains}

\author{
Zhaopei Huang$^1$
\and
Jinming Zhao$^{2\ast}$ \and
Qin Jin$^{1\ast}$ 
\affiliations
$^1$Renmin University of China\\
$^2$Independent Researcher
\emails
\{huangzhaopei, qjin\}@ruc.edu.cn,
zhaojinming1@gmail.com
}

\begin{document}

\maketitle

\renewcommand{\thefootnote}{\fnsymbol{footnote}} 
\footnotetext[1]{Qin Jin and Jinming Zhao are corresponding authors.}
\renewcommand{\thefootnote}{\arabic{footnote}} 
\setcounter{footnote}{0}  

\begin{abstract}

Understanding the process of emotion generation is crucial for analyzing the causes behind emotions. Causal Emotion Entailment (CEE), an emotion-understanding task, aims to identify the causal utterances in a conversation that stimulate the emotions expressed in a target utterance.
However, current works in CEE mainly focus on modeling semantic and emotional interactions in conversations, neglecting the exploration of the emotion-generation process. 
This hinders the models from deeply understanding emotions, restricting their ability to produce explainable predictions.
In this work, inspired by the emotion generation process of ``{\it stimulus-appraisal-emotion}" in the cognitive appraisal theory, we introduce a step-by-step reasoning method, \textbf{E}motion-\textbf{C}ause \textbf{R}easoning \textbf{Chain} (\textbf{ECR-Chain}), to infer the stimulus from the target emotional expressions in conversations.
Specifically, we first introduce the ECR-Chain to ChatGPT via few-shot prompting, which significantly improves its performance on the CEE task. 
We further propose an automated construction process to utilize ChatGPT in building an \textbf{ECR-Chain set}, which can enhance the reasoning abilities of smaller models through supervised training and assist the Vicuna-7B model in achieving state-of-the-art CEE performance.
Moreover, our methods can enable these generative language models to effectively perform emotion-cause reasoning in an explainable manner.
Our code, data and more details are at \href{https://github.com/hzp3517/ECR-Chain}{https:// github.com/hzp3517/ECR-Chain}.
\end{abstract}

\section{Introduction}
\label{sec:intro}
Emotion plays an important role in human communication and there are many dialogue-related research works in the NLP field involving analysis and utilization of emotions. Emotion recognition in conversation, for example, aims to detect the emotional status of a conversation utterance based on the context~\cite{majumder2019dialoguernn,ghosal2019dialoguegcn,liu2022dialogueein}. Emotional response generation, on the other hand, aims to incorporate emotional factors into dialog systems to enhance user satisfaction~\cite{wei2019emotion_response_generation,zhong2021care}. 
However, these tasks lack attention to deeply understanding the causes behind emotions, which limits the application in broader areas like emotional support dialogue systems.

\begin{figure}[t]
\centerline{\includegraphics[width=1.0\columnwidth]{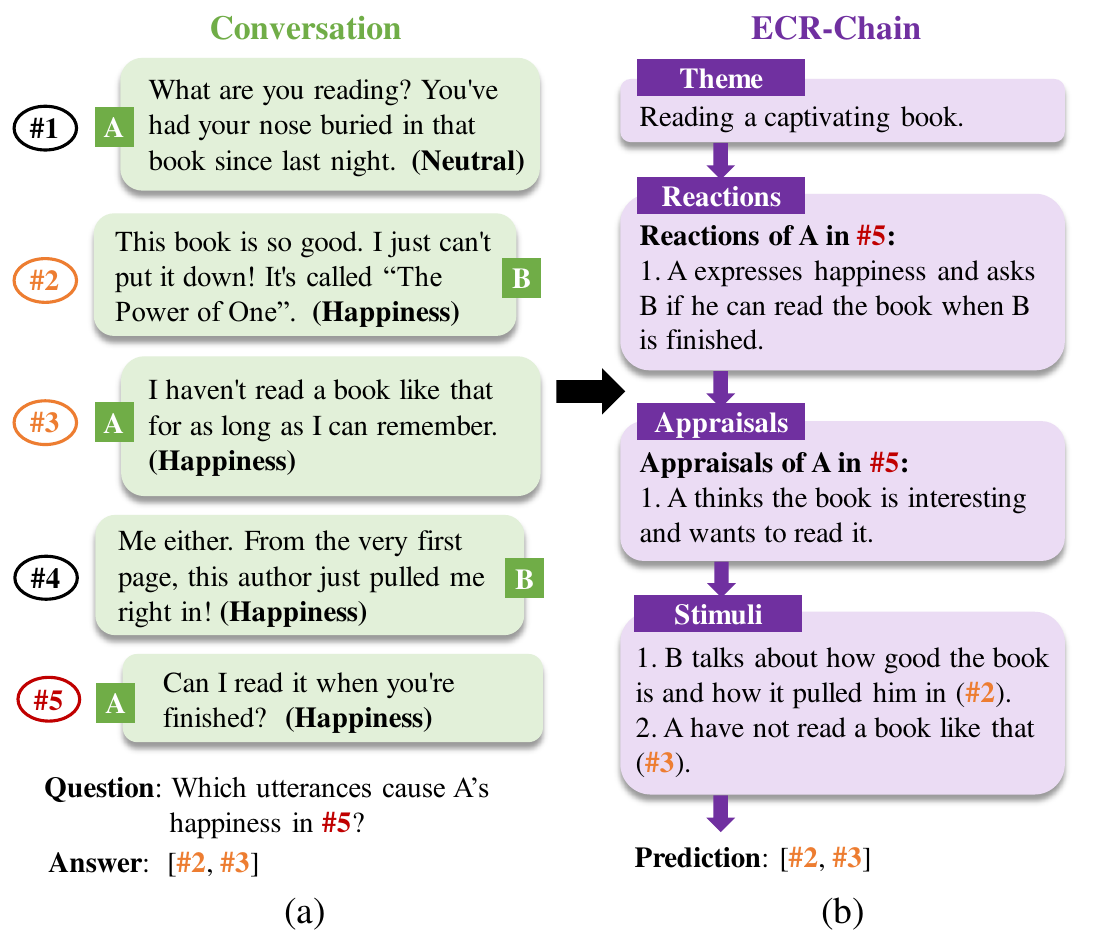}}
\vspace{-6pt}
\caption{An example of emotion-cause reasoning in a conversation. \textbf{(a)}: The basic form of the CEE task, requires identifying the causal utterances (\#2 \& \#3) that stimulate the speaker's emotion in the target utterance (\#5). \textbf{(b)}: Our proposed reasoning method, ECR-Chain, begins by summarizing the conversation theme, then describes the reactions and infers the appraisals of the target speaker, and finally deduces emotion-causes.
}
\label{fig:ecr_chain}
\vspace{-2mm}
\end{figure}

Therefore, a new task called ``Causal Emotion Entailment" (CEE) together with a new benchmark dataset ``RECCON"~\cite{poria2021reccon} has been proposed. As depicted in Figure~\ref{fig:ecr_chain}, 
given a two-person conversation, the corresponding emotional state of the speaker of each sentence, and the target utterance,
the goal of CEE is to identify which preceding utterances (including the target itself) are causal utterances containing the stimulus factors that lead to the emotion of the target utterance. 
Based on this benchmark, various methods that attempt to improve emotion-cause prediction have been explored.
Modeling the speaker-aware semantic and emotional interaction in the conversation context has been proven to be effective~\cite{zhang2022tsam} while introducing external social commonsense knowledge (CSK)~\cite{hwang2021comet} to the contextual interaction could also be helpful~\cite{li2022kec,zhao2023kbcin}.
These methods treat the CEE task as a supervised classification problem, identifying whether each utterance is a causal utterance or not.
Nevertheless, they neglect the exploration and learning of the emotion-generation process, which limits their ability to take a deeper understanding of emotions and restricts them from providing explainable rationales behind the emotion-cause.
In contrast to these aforementioned works, we will not only explore to better address the CEE task but also extend it into a more challenging Explainable Emotion-Cause Reasoning task in this paper.

To overcome the limitations of previous models and realize explainable emotion-cause reasoning, we explore using generative language models to reason in our task. Considering that inferring the causes from emotional expression can be seen as a reverse process of emotion generation, we can refer to psychological descriptions of how emotions arise to determine the reasoning steps. According to the cognitive appraisal theory~\cite{arnold1960emotion,ellsworth1991some}, the process of emotion generation can be summarized as ``{\it stimulus-appraisal-emotion}". This sequence starts when a person attends to certain aspects of their environment or events, stimulating them to form interpretations or evaluations internally, which ultimately culminate in a specific emotional response. Therefore, in our task, considering our textual conversation scenarios, we can start by identifying behaviors related to the target emotion based on the target utterance and the global context. Subsequently, we can infer the speaker's inner thoughts based on their behaviors and finally deduce the stimuli that led to these thoughts. Based on the above analysis, we introduce an \textbf{E}motion-\textbf{C}ause \textbf{R}easoning \textbf{Chain} (\textbf{ECR-Chain}), formatted as ``{\it theme} $\to$ {\it reaction} $\to$ {\it appraisal} $\to$ {\it stimulus}", which can be regarded as a step-by-step reasoning process as shown in Figure~\ref{fig:ecr_chain}.

Specifically, inspired by the Chain-of-Thought (CoT) prompting method~\cite{wei2022cot}, we incorporate our ECR-Chain into the in-context learning~\cite{brown2020language} to guide large language model's reasoning step by step and achieve significant performance improvements under the few-shot CEE task setting. Having verified the effectiveness of this chain in emotion-cause reasoning, we further supplement the RECCON dataset with an automatically generated ECR-Chain set that can be used for supervised training. For example, we use the constructed ECR-Chain set for multi-task training on smaller generative language models and effectively enhance their emotional reasoning abilities to provide explainable reasoning paths and cause descriptions that were difficult to achieve with previous CEE models.

The main contributions of this work include:
1) We propose an emotion-cause reasoning method, ECR-Chain, and use it to guide the reasoning process of language models via CoT prompting. 2) We automatically construct an ECR-Chain set to help enhance the emotional reasoning ability of supervised models. 3) Extensive experimental results over various settings demonstrate the effectiveness of our method for predicting emotion-cause utterances and performing explainable emotion-cause reasoning.

\section{Related Works}
\label{sec:realted_work}
\subsection{Causal Emotion Entailment}
\label{ssec:emo_cause_recognition}
Poria et al.~\shortcite{poria2021reccon} address the task of recognizing emotion-causes in conversations. It includes two sub-tasks: Causal Emotion Entailment (CEE) and Causal Span Extraction (CSE). Recent related works mostly focus on the CEE task. 
Zhang et al.~\shortcite{zhang2022tsam} propose a two-stream attention model (TSAM) to incorporate both speaker identities and emotional states into utterance features via contextual interaction modeling.
Li et al.~\shortcite{li2022kec} introduce commonsense knowledge (CSK) related to speakers' emotional interactions and build a knowledge-enhanced conversation graph. Zhao et al.~\shortcite{zhao2023kbcin} further leverage event-centered CSK and social-interaction CSK to construct a knowledge-bridged causal interaction network.
These above works all use RoBERTa~\cite{liu2019roberta} to extract contextual utterance-level features and design specific graph structures to model various interactions between utterances. Although their model structures are suited for making classifications for each utterance in a conversation, the explainability of their models is limited.
Recently, Zhao et al.~\shortcite{zhao2023chatgpt_emotion} evaluate the performance of ChatGPT~\footnote{https://openai.com/blog/chatgpt} on various emotion-related dialogue tasks, including the CEE.
However, they simply instruct the ChatGPT to provide predictions based on the task input, lacking exploration of reasoning methods, which limits their performance.
In our work, we explore employing a reasoning process to enhance the emotion-cause prediction performance of generative language models and enable them to have explainable reasoning capabilities.

\subsection{Chain-of-Thought for Reasoning}
\label{ssec:cot_for_reason}
Wei et al.~\shortcite{wei2022cot} first explore adding a few Chain-of-Thought (CoT) demonstrations via few-shot prompting to guide the large language model (LLM) thinking ``step-by-step". Following their success, lots of works continue exploring improving the CoT prompting methods, including Self-Consistency~\cite{wang2022sc}, DiVeRSe~\cite{li2022diverse}, Auto-CoT~\cite{zhang2022autocot}, and Active-Prompt~\cite{diao2023active}. On the other hand, though smaller models cannot directly benefit from CoT prompting~\cite{wei2022cot}, distilling the rationales generated by LLM into smaller models may still be helpful for them to solve the reasoning tasks~\cite{ho-etal-2023-finetunecot}. 
Li et al.~\shortcite{li2022mtcot} first prompt LLM to generate rationales for each question in the dataset. 
They then propose a multi-task training strategy to utilize the rationales as auxiliary training objectives for smaller models. Hsieh et al.~\shortcite{hsieh-etal-2023-distilling} also apply the multi-task method to distill the reasoning knowledge and improve the smaller models to outperform LLMs on their tasks. Inspired by these works, we first propose a CoT method customized for emotion-cause reasoning, and then improve the smaller models via multi-task training with the rationales automatically constructed by LLMs.

\section{Method}
\label{sec:method}
\subsection{Task Definition}
\label{ssec:task_definition}
For the {\it Causal Emotion Entailment (CEE)} task, given a two-party conversation $C$, which is denoted as $C = [(u_1, k_1, e_1), (u_2, k_2, e_2), ..., (u_t, k_t, e_t)]$, where $k_i$, $u_i$ and $e_i$ are the speaker identity, the content, and the emotion of the $i$-th utterance respectively, its goal is to predict which utterances $u_i \  (i \leq t)$ in the conversation context are responsible for evoking the non-neutral emotion $e_t$ of the target utterance $u_t$. We can reform it into a question-answer task, where the question is $Q=\{C, (u_t, k_t, e_t)\}$ and the answer is $A=\{u_i\}$. CEE task only evaluates the predicted answer $A$ and does not care how the answer is derived. 

To develop explainable models, we further extend the CEE task to an {\it Explainable Emotion-Cause Reasoning (Explainable ECR)} task, where given a question $Q$ in the same format as mentioned above, the model is required to output the predicted answer $A$ along with its rationale $R$. Explainable CEE may enhance the credibility of the prediction and benefit other emotion-related tasks.

\subsection{Overview}
\label{ssec:overview}
Based on the cognitive appraisal theory in psychology, we introduce a reasoning process of ``{\it theme} $\to$ {\it reaction} $\to$ {\it appraisal} $\to$ {\it stimulus}", which we call it ``\textbf{E}motion-\textbf{C}ause \textbf{R}easoning \textbf{Chain} (\textbf{ECR-Chain})". Tailored for emotion-cause reasoning in textual conversations, this serialized reasoning scheme is also consistent with the generation mode of generative language models, as they always predict the following words based on the preceding content. Therefore, we consider taking advantage of the ECR-Chain to improve the emotion-cause reasoning of generative language models. 

For the two tasks defined in Section~\ref{ssec:task_definition}, we explore both the few-shot learning with large language models (LLMs) and the supervised learning with smaller generative language models~\footnote{Models with a size smaller than 100B are considered to lack evident CoT ability~\cite{wei2022cot}.}. 
For LLMs larger than 100B, inspired by the Chain-of-Thought (CoT) method~\cite{wei2022cot}, we design a few-shot prompt to instruct them to reason step-by-step following our proposed ECR-Chain and then infer the final answer. But for smaller models, due to their limited parameter scales, the effect of directly applying the CoT prompting is not evident~\cite{wei2022cot}. So we consider improving their reasoning abilities through supervised training. As no such ECR-Chain data is available, we employ LLMs to automatically construct an ECR-Chain set to enable supervised training for smaller models to enhance their reasoning capability. Ultimately, through multi-task training, we build a model capable of performing the two tasks.

\subsection{Few-shot Prompting with ECR-Chain}
\label{ssec:fewshot_method}
For generative language models, the most intuitive solution to a question-answer task is to generate an answer $A$ directly based on the question $Q$, that is, $P(A|Q)$. However, when performing a reasoning task, the indirect correlation between $Q$ and $A$ may limit the performance. For LLMs, we can employ the CoT prompting method to first guide them to generate a rationale $R$ for the question $Q$, and then derive the final answer $A$ based on $R$, forming in the format $P(R,A|Q)$. This approach often results in more accurate predictions. 
Considering that inferring emotion causes from emotional expressions also requires step-by-step reasoning, we design a few-shot prompt, denoted as \textless reasoning\textgreater \ ,  to guide the LLMs in reasoning along the ECR-Chain. Figure~\ref{fig:cot_prompt} is a diagram of our prompt, which mainly contains four parts:

\begin{figure}[t]
\begin{prompt_box}
\small 
Please understand the emotion-cause for the target utterance in a given conversation.\\
\textbf{[Task Description]}: ...\\
\textbf{[Description of Reasoning Process]}:\\
\sethlcolor{yellow} 
1. Output the ``\hl{Theme}": ...\\
2. List ``\hl{Reactions}" items of target utterance: ...\\
3. List ``\hl{Appraisals}" items of target utterance: ...\\
4. List ``\hl{Stimuli}" items with their corresponding utterance id: ...\\
5. Based on these ``Stimuli" items, output the index number of the \hl{causal utterances} in the form of Python list without any other content.\\
\\
I will show you some examples:\\
\textbf{[Example]}: ...\\
\\
--- To be solved ---\\
\textbf{[Question]}: ...
\end{prompt_box}
\vspace{-8pt}
\caption{Illustration of our designed few-shot prompt for reasoning along the ECR-Chain. Detailed prompt examples are presented in the Appendix.}
\label{fig:cot_prompt}
\vspace{-2mm}
\end{figure}

\begin{itemize}[leftmargin=*]
\item \textbf{Task Description}: Contains the goals of the emotion-cause reasoning task and the input and output formats.
\item \textbf{Description of Reasoning Process}: Describe the meaning of each component in the ECR-Chain and specific requirements for each of the 5-step reasoning steps. We instruct the model to list the ``{\it Reactions}", ``{\it Appraisals}", ``{\it Stimuli}" in items to make the reasoning process clearer. Besides, we require that each generated ``{\it Stimulus}" item be followed by its corresponding utterance indexes so that the model can directly summarize all predicted utterance indexes and provide the final answer.
\item \textbf{Example}: We manually write expected rationales for a few of the questions in the training set to serve as exemplars in few-shot prompting. Each exemplar consists of the conversation $C$, the target utterance $u_t$, and a manually written rationale $r$. Each utterance with its emotion and speaker label is simply written in this format: \#[utterance index]: [speaker] ([emotion]): ``[utterance content]".
\item \textbf{Question}: Present the conversation and target utterance for each sample in the test data and instruct the model to perform 5-step reasoning.
\end{itemize}

\subsection{Supervised Learning with ECR-Chain}
\label{ssec:supervised_method}

\begin{figure*}[t]
\centerline{\includegraphics[width=0.95\textwidth]{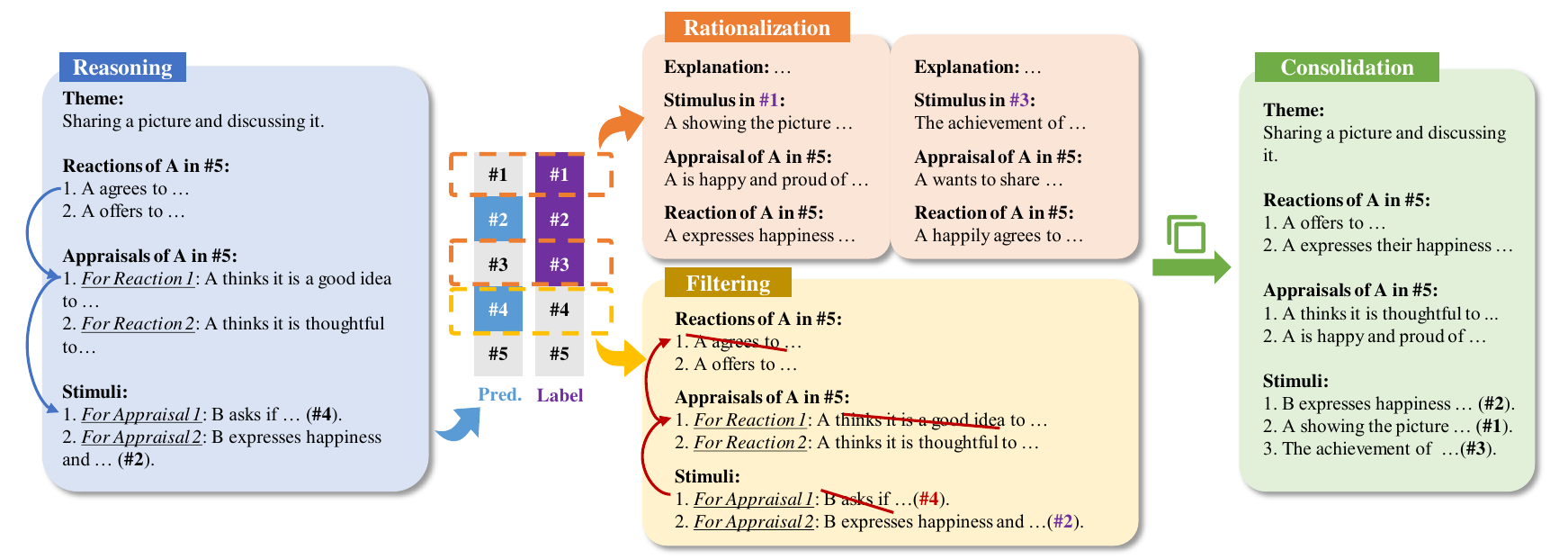}}
\vspace{-6pt}
\caption{An example of the automated construction process of ECR-Chain set. \textbf{1) Reasoning}: Instruct the LLM with a \textless reasoning\textgreater \  prompt to generate a raw rationale without providing labels; \textbf{2) Filtering}: Delete chains which lead to incorrect answers (\#4); \textbf{3) Rationalization}: For each golden answers not predicted (\#1 \& \#3), supplement an additional chain by instructing the LLM with a \textless rationalization\textgreater \  prompt; \textbf{4) Consolidation}: Merge all retained and supplemented chains and consolidate semantically similar items by instructing the LLM with a \textless consolidation\textgreater \  prompt.
}
\label{fig:chain_construct}
\vspace{-2mm}
\end{figure*}

Since the reasoning capability of smaller models is limited, it is hard to teach them to reason along our introduced ECR-Chain simply through few-shot prompting. However, recent studies have shown that distilling the CoT rationales from LLM to smaller ones can help to enhance the reasoning capabilities of the smaller models on specific tasks~\cite{li2023dissecting,ho-etal-2023-finetunecot}. To this end, we utilize LLMs to automatically construct an ECR-Chain set, and perform supervised training with ECR-Chain on smaller models.

\subsubsection{Automated Construction of ECR-Chain Set}
\label{sssec:chain_construction}

The generation of reasoning chains can be approached in two ways: 1) reasoning through CoT prompting which derives an answer, or 2) generating explanations by rationalization prompting conditioned on labels~\cite{marasovic2022self_rationalization}. The first approach may generate higher-quality rationales if the derived prediction is correct since the rationales have been proven to effectively guide the language model to output our expected answer. However, when the prediction is wrong, the reasoning approach may provide some misleading rationales. 
Zelikman et al.~\shortcite{zelikman2022star} and Li et al.~\shortcite{li2022mtcot} propose to generate a new rationale by rationalization prompting to supplement for questions that can not be reasoned to a correct answer. However, our ECR-Chain rationales are more complex compared to their works, as our rationales may contain multiple items in the {\it Reaction}, {\it Appraisal}, and {\it Stimulus} part, and different stimulus items may lead to different causal utterance predictions. Therefore, we propose a 4-stage processing procedure to generate and further revise our rationales, which involves reasoning, filtering, rationalization, and consolidation, as illustrated in Figure~\ref{fig:chain_construct}.

\textbf{a) Reasoning:}
We first execute a reasoning process using the \textless reasoning\textgreater \ prompt illustrated in Figure~\ref{fig:cot_prompt} and obtain raw rationales. Here we denote the {\it Reaction}, {\it Appraisal}, and {\it Stimulus} part of our rationals as $\boldsymbol{n}$, $\boldsymbol{l}$, $\boldsymbol{s}$, respectively. Each of them may contain multiple items. Here, we require that for each appraisal item $l_j$ generated, the index number of its corresponding reaction item $n_i$ be specified. For each stimulus item $s_k$, its corresponding $l_j$ needs to be specified similarly. From this, we can determine the corresponding relationship from $n_i$ to $l_j$ and to $s_k$, linking these three items to a chain, as shown in Figure~\ref{fig:chain_construct}. Besides, we can find the set of all predicted answers $\boldsymbol{a}^{pred}$ corresponding to generated stimulus items $\boldsymbol{s}$.

\textbf{b) Filtering:}
Since the ground-truth label is unknown to the LLM in the reasoning stage, the raw rationales may include some chains that lead to wrong answers. With the help of the connections we built in the reasoning stage, we can remove an entire chain $\{n_i, l_j, s_k\}$ corresponding to a wrong predicted answer $a^{pred}_w$. We also remove those items in each part that could not lead to any subsequent items.

\textbf{c) Rationalization:}
If there are some answers in the ground-truth label that do not appear in $\boldsymbol{a}^{pred}$, we then need to supplement their corresponding rationales. We perform a rationalization process separately for each missing labeled answer $a^{GT}_m$. Intuitively, rationalization should be the opposite process of reasoning, that is, deducing the target speaker's appraisal of the stimulus factors corresponding to the specified causal utterance, and then further deriving the corresponding reaction. However, extracting the exact stimulus factors from a given causal utterance may not be as straightforward as locating the causal utterance based on already inferred stimulus factors. Because the stimulus factors are often only a part of the full semantic content of its corresponding causal utterance. Therefore, it is necessary to consider the entire process of the emotion generation process first. In our designed \textless rationalization\textgreater \ prompt, we first add an ``explanation" step. In this step, we ask the LLM to explain in free text why the specified causal utterance would trigger the target speaker's target emotion, and what the evoking process is like, to fully consider the connection between the target utterance and the specified causal utterance. Following this, the model can extract a stimulus factor from the causal utterance more precisely and then deduce its subsequent appraisal and response. This approach results in clearer and more precise rationales, especially in the stimulus part.

\textbf{d) Consolidation:}
For a conversation sample, we need to merge the reasoning chains after filtering and rationalization.
However, the merged rationale may contain many semantically similar items in each part. To make it more concise and clear, we design a \textless consolidation\textgreater \  prompt to instruct the LLM to condense the items in the {\it Reaction}, {\it Appraisal}, and {\it Stimulus} parts if needed. We require the LLM to merge items with similar semantics in each part while avoiding excessive modifications to the original phrasing.

\subsubsection{Multi-task Training with ECR-Chain Set}
\label{sssec:multitask}
After constructing the ECR-Chain set, we can utilize it to assist the supervised training of smaller models. We adopt the MT-CoT method proposed by~\cite{li2022mtcot} to train on both the golden answer and the collected rationales. Specifically, we utilize two types of instructions. The one uses a simple \textless answer\textgreater \ prompt that guides the model to directly output the answer based on the question, namely $P(A|Q)$. And the other applies the \textless reasoning\textgreater \ prompt to instruct the model performing $P(R,A|Q)$. During the training process, we construct each question sample into these two types of instructional inputs separately and calculate the ${\cal L}_{reasoning}$ and ${\cal L}_{answer}$ with the referenced rationale constructed by LLM and the golden causal utterance label, respectively. The overall loss function is simply a mixture of these two losses:
\begin{equation}
    {\cal L}_{MT} = {\cal L}_{reasoning} + {\cal L}_{answer}
\end{equation}

This approach enables our trained model to perform inference in different manners. If we only need to identify the causal utterances in applications like the CEE task, we can use the \textless answer\textgreater \ instruction to predict directly. If we expect an explainable generation, we can use the \textless reasoning\textgreater \ instruction to provide step-by-step descriptions of the reasoning process.

\section{Experiment}
\label{sec:experiment}
\subsection{Dataset}
\label{ssec:dataset}

\begin{table}[t]\small
\centering
\begin{tabular}{lrrr}
\toprule
Statistics               & Train & Valid & Test \\ \midrule
Positive Causal Pairs    & 7,027  & 328   & 1,767 \\
Negative Causal Pairs    & 20,646 & 838   & 5,330 \\
Unique Conversations         & 834   & 47    & 225  \\
Samples                  & 4,562  & 200   & 1,099 \\ \bottomrule
\end{tabular}
\caption{Dataset statistics. We consider each target utterance as a sample. A conversation may contain several target utterances, forming several samples.}
\label{tab:statistics}
\vspace{-2mm}
\end{table}

We conduct experiments on the RECCON-DD dataset~\cite{poria2021reccon}. This dataset supplements causal utterance annotations for each non-neutral utterance in the conversations of the DailyDialog dataset~\cite{li2017dailydialog}. For a target utterance, its corresponding causal utterances form positive causal pairs with it, while the remaining utterances in the conversation history form negative pairs with it. Following~\cite{zhao2023kbcin}, we only consider the potential causes within the conversation history and remove repetitive positive causal pairs in the original dataset.
The statistical details of the dataset are presented in Table~\ref{tab:statistics}. Following Poria et al.~\shortcite{poria2021reccon}, we report the F1 scores of both negative and positive causal pairs and the macro F1 scores of them when evaluating performance on the CEE task.

\subsection{Implementation Details}
\label{ssec:implementation}
We utilize ChatGPT (gpt-3.5-turbo-0613) as our LLM and set the temperature to 0 to reduce the randomness. During the construction of the ECR-Chain set, we employed 4 hand-written exemplars for the \textless reasoning\textgreater \ prompt, 4 exemplars for the \textless rationalization\textgreater \ prompt, and 3 for the \textless consolidation\textgreater \ prompt to enhance the LLM's understanding of our specific requirements. 
The details about the examplars and prompts are presented in the Appendix. 
For the smaller language model, we opt for Vicuna-7B-v1.3~\footnote{https://lmsys.org/blog/2023-03-30-vicuna/} which is based on the LLaMA~\cite{touvron2023llama}. Due to the computation limit, we applied LoRA fine-tuning~\cite{hu2021lora} for the supervised training. Our total training batch is set to 256 (with gradient accumulation) and the learning rate is set to 1e-3. We train 10 epochs and pick the model that performed best on the validation set to evaluate on the test set. We report the average results of 3 runs for the supervised model.

\subsection{CEE with Few-shot Learning}
\label{ssec:CEE_fewshot}

\begin{table}[t]\small
\centering
\begin{tabular}{llccc}
\toprule
 & Method & Neg. F1 & Pos. F1 & Macro F1 \\ \midrule
\multicolumn{1}{c}{} & Zhao et al.$^{\dagger}$ & 82.10 & 52.84 & 67.47 \\
\multicolumn{1}{c}{} & Answer & 80.41 & 47.00 & 63.71 \\
\multicolumn{1}{c}{\multirow{-3}{*}{1-shot}} & Reasoning & \textbf{85.92} & \textbf{55.05} & \textbf{70.48} \\ \midrule
 & Answer & 77.40 & 53.46 & 65.43 \\
\multirow{-2}{*}{4-shot} & Reasoning & \textbf{87.40} & \textbf{58.97} & \textbf{73.19} \\ \bottomrule
\end{tabular}
\caption{Results of Few-shot CEE. $^{\dagger}$ denotes the results referred from~\protect\cite{zhao2023chatgpt_emotion}. ``Answer" means instructing the LLM to directly predict the answer, whereas ``Reasoning" means guiding the LLM to follow the ECR-Chain to derive the answer.}
\label{tab:unsupervised_main}
\end{table}

\begin{table}[t]\small
\centering
\begin{tabular}{llllccc}
\toprule
T & R & A & S & Neg. F1 & Pos. F1 & Macro F1 \\ \midrule
$\checkmark$ & $\checkmark$ & $\checkmark$ & $\checkmark$ & \textbf{87.40} & \textbf{58.97} & \textbf{73.19} \\
$\times$ & $\checkmark$ & $\checkmark$ & $\checkmark$ & 87.35 & 57.71 & 72.53 \\
$\checkmark$ & $\times$ & $\checkmark$ & $\checkmark$ & 85.54 & 55.32 & 70.43 \\
$\times$ & $\times$ & $\checkmark$ & $\checkmark$ & 85.93 & 58.00 & 71.97 \\
$\times$ & $\times$ & $\times$ & $\checkmark$ & 78.28 & 47.31 & 62.79 \\ \bottomrule
\end{tabular}
\caption{Ablation study on the individual parts of the ECR-Chain. ``T", ``R", ``A", and ``S" correspond to theme, reaction, appraisal, and stimulus, respectively.}
\label{tab:unsupervised_ablation}
\vspace{-2mm}
\end{table}

We first report the results of using LLMs to perform the CEE task in a few-shot manner, as shown in Table~\ref{tab:unsupervised_main}. Zhao et al.~\cite{zhao2023chatgpt_emotion} have evaluated the performance of ChatGPT on various emotion-related tasks, including the CEE task. They construct task-specific prompts and ask the model to predict the answer without attempting to guide the model through reasoning. Since they do not provide prompt details in their paper, we directly reference the results they reported. We then compare the CEE performance of directly predicting the answers with reasoning through our ECR-Chain, both under our input format. In the 1-shot scenario, the performance of ``Answer" in our experiments is inferior to that of~\cite{zhao2023chatgpt_emotion}, which might be due to discrepancies in input format, ChatGPT version, or the chosen exemplar. But our primary focus here is on the effectiveness of reasoning using the ECR-Chain. According to Table~\ref{tab:unsupervised_main}, guiding the LLM to reason following the ECR-Chain significantly improves causal utterance prediction over the two aforementioned baselines. In the 4-shot scenario, our proposed reasoning method can bring more than a 7\% improvement in the Macro F1 metric over the baseline, which demonstrates the benefit of the reasoning process in correctly identifying emotion-causes.

We also conduct an ablation study to analyze the impact of each part of our proposed reasoning chain, as shown in Table~\ref{tab:unsupervised_ablation}. The stimulus part, which is directly associated with the final causal utterances, is considered a detailed description of the cause and thus was retained in all experiments to assess the influence of the presence or absence of other components. When the LLM is instructed to directly provide a stimulus description based on the input, the predicted causal utterances corresponding to the stimulus are often inaccurate. In contrast, the accuracy of tracing the cause improves significantly if the model first deduces the speaker's appraisal associated with the target emotion before inferring the stimulus that evoked that emotion. This indicates that ``appraisal" is central to the entire process of emotion generation, and understanding appraisal is crucial for comprehending emotions. Additionally, summarizing the conversation's theme and the speaker's emotional reaction is beneficial. It helps the LLM gain a fuller understanding of the dialogue context and the speaker's state, thereby enhancing the causal tracing.

\subsection{CEE with Supervised Learning}
\label{ssec:supervised_CEE}

\begin{table}[t]\small
\centering
\begin{tabular}{lcccc}
\toprule
Method & Expl. & Neg. F1 & Pos. F1 & Macro F1 \\ \midrule
RoBERTa-B & $\times$ & 88.74 & 64.28 & 76.51 \\
RoBERTa-L & $\times$ & 87.89 & 66.23 & 77.06 \\
KEC$^{\dagger}$ & $\times$ & 88.85 & 66.55 & 77.70 \\
KBCIN & $\times$ & 89.65 & 68.59 & 79.12 \\
TSAM & $\times$ & 90.48 & 70.00 & 80.24 \\ \midrule
Answer & $\times$ & 89.78 & 68.66 & 79.22 \\
Reasoning & $\checkmark$ & 89.70 & 63.36 & 76.53 \\ \hdashline
\multirow{2}{*}{Multi-task} & $\times$ & \textbf{90.82} & \textbf{70.84} & \textbf{80.83} \\
 & $\checkmark$ & 90.33 & 65.89 & 78.11 \\ \bottomrule
\end{tabular}
\caption{Results of Supervised CEE. ``Expl." indicates whether the answer is predicted in an explainable manner. Our multi-task trained model is capable of both directly predicting answers and providing explainable rationales during inference, resulting in two separate lines in this table. $^{\dagger}$ denotes that we reference the results from ~\protect\cite{zhao2023kbcin}, as the original KEC paper treated neutral utterance as target utterance as well, which differs from our setting.}
\label{tab:supervised}
\vspace{-2mm}
\end{table}

As shown in Table~\ref{tab:supervised}, we compare our multi-task model against models trained on single tasks, namely the ``Answer" and ``Reasoning" listed in the table. ``Answer" refers to only employing the CEE task labels for training the Vicuna. During inference, the model directly outputs the predicted answers, which are not explainable. ``Reasoning" denotes fully supervising the Vicuna by ECR-Chains generated by the LLM during training. The trained model outputs rationales in a step-by-step manner during inference, making the results explainable. The multi-task model, as described in Section~\ref{sssec:multitask}, is trained using both answer-only and reasoning supervision, allowing the model to either directly predict answers or generate rationales to derive the answer, depending on the form of the input instruction.
Besides, we report the results of some previous methods based on RoBERTa: \textbf{KEC}~\cite{li2022kec}, \textbf{KBCIN}~\cite{zhao2023kbcin}, \textbf{TSAM}~\cite{zhang2022tsam}, along with the results of vanilla RoBERTa. These previous works extract features from utterances, model interactions between different features within the conversation, and then perform binary classification on each utterance feature to obtain a prediction of either positive or negative. Hence, the predictions from these works are unexplainable. 

Compared to the answer-only training strategy, incorporating the ECR-Chain set for multi-task training can bring an obvious performance gain when utilizing the \textless answer\textgreater \  instruction for inference. Our multi-task model, when directly predicting answers, also surpasses the previous baselines. This indicates that our ECR-Chain set, providing more detailed and comprehensive supervision, can improve the supervised models to capture emotion-causes more precisely. 

When comparing the results of ``Answer" and ``Reasoning", we observe that performing reasoning in training and inference is less effective than directly predicting the answer for the CEE task. We think there are two possible reasons. Firstly, the long length of our rationales may lead to insufficient optimization of the model for the final answer prediction part, which constitutes only a small portion of an entire rationale. Secondly, the reasoning ability of the smaller model may be limited by its scale. We consider this an open problem, as similar phenomena were also observed in the one-stage fine-tuning experiments of~\cite{zhang2023multimodalcot}. The multi-task training method can mitigate the first issue, and our experiments show that the predictions derived through reasoning in multi-task models indeed bring improvement. Moreover, the explainable outcomes achieved by our multi-task model via \textless reasoning\textgreater \ instruction are comparable to those unexplainable baselines specifically training for the CEE task from previous works, further highlighting the advantages of our model.

In summary, our multi-task model can be regarded as a versatile model, capable of providing precise predictions directly via the \textless answer\textgreater \  instruction, as well as offering explainable reasoning of the predictions through the \textless reasoning\textgreater \  instruction.

\subsection{Explainable Emotion-Cause Reasoning}
\label{ssec:explainable_ECR}

\begin{table}[t]\small
\centering
\begin{tabular}{lccc}
\toprule
Method & Macro F1 & GPT4 Score & Claude3 Score \\ \midrule
Vanilla-Vicuna & 53.41 & 6.63 & 7.22 \\
MT-Vicuna & \textbf{78.11} & \ul{7.68} & \ul{7.56} \\
ChatGPT & \ul{70.48} & \textbf{8.16} & \textbf{7.99} \\ \bottomrule
\end{tabular}
\caption{Results of Explainable ECR. The ``GPT4 Score" and ``Claude3 Score" evaluates the quality of rationales, with the score ranging from 1 to 10.}
\label{tab:explainable_ecr}
\vspace{-2mm}
\end{table}

\begin{figure*}[t]
\centerline{\includegraphics[width=1.0\textwidth]{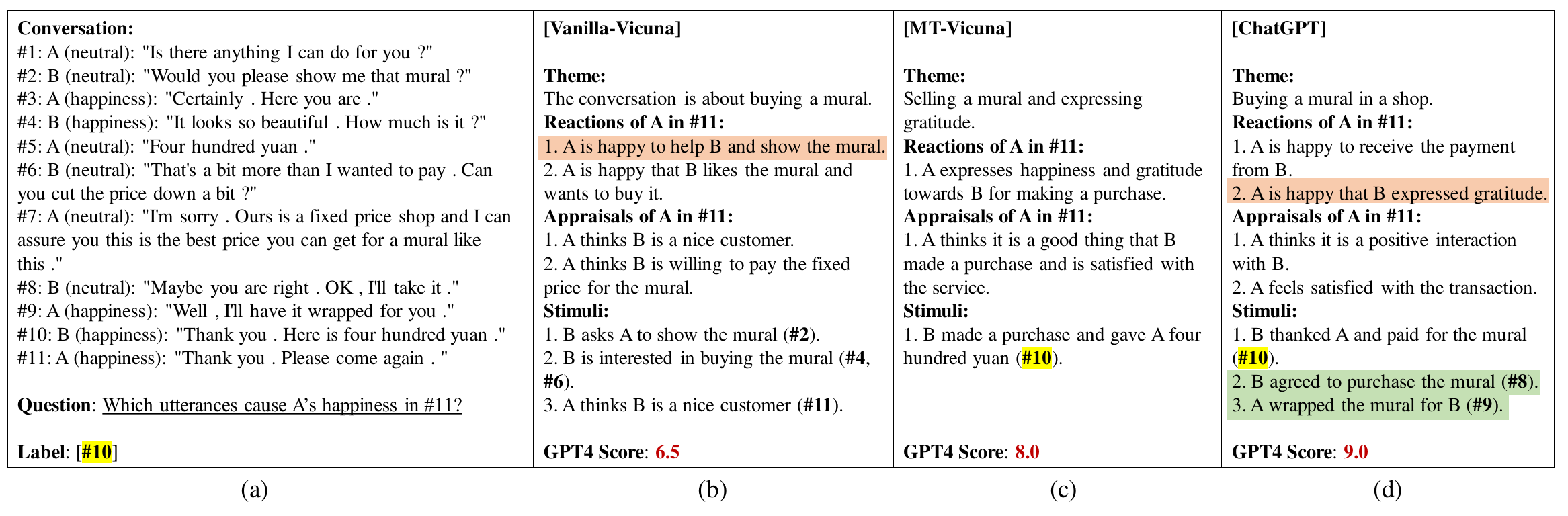}}
\vspace{-6pt}
\caption{Case Study on Explainable ECR. (a) shows the question and the labeled causal utterances. (b), (c), and (d) show rationales generated by three models. According to the analysis of GPT-4, the output of our multi-task model is considered to be ``more concise and straightforward". The other two models, however, are thought to ``contain inaccuracies or assumptions", as indicated by the orange background content in (b) and (d). Nonetheless, GPT-4 states that ChatGPT ``provides a more detailed analysis of the stimuli" and ``exhibits a more comprehensive understanding of the situation", as indicated by the green background content in (d). This is the reason for the higher score awarded to ChatGPT.}
\label{fig:case_study}
\vspace{-2mm}
\end{figure*}

Here we evaluate the performance of models when performing emotion-cause reasoning in an explainable manner, as shown in Table~\ref{tab:explainable_ecr}. ``Vanilla-Vicuna" refers to the original Vicuna model without our fine-tuning, while ``MT-Vicuna" is our multi-task trained model. All models apply the same 1-shot \textless reasoning\textgreater \  instruction as input. ``Macro F1" evaluates the final reasoned causal utterance indexes against golden labels. The ``GPT-4/Claude3 Score", on the other hand, employs advanced LLMs, GPT-4 and Claude3-Opus, to assess the quality of the generated rationales. Specifically, we follow the evaluation template provided in FairEval~\cite{wang2023faireval} to design our own evaluating prompt. We instruct it to focus on three aspects of the rationale: logical coherence within the rationale, the relevance of the rationale to the original conversation, and the plausibility of the final stimulus description. It is required to give an overall score for each rationale, together with the scoring explanations. More detailed evaluation settings and evaluation prompts are presented in the Appendix.

For the ``Macro F1" score of causal utterance prediction in Table~\ref{tab:explainable_ecr}, our multi-task model performs the best, even surpassing the results of the teacher model ChatGPT, which indicates that our ECR-Chain set refined based on the golden labels can effectively aid supervised models in deriving the expected answers. And for the metric of rationale quality, ``GPT-4/Claude3 Score", our multi-task model also significantly outperforms the vanilla vicuna, proving that our constructed ECR-Chain set can effectively enhance the reasoning abilities and the explainability for emotion causes via supervised training. However, there is still a certain gap between our multi-task model and ChatGPT. On the one hand, this may be due to ChatGPT providing rationales richer in logical coherence and more aligned with the conversation context, which reflects the powerful logical and explanatory capabilities of ChatGPT brought about by its large parameter scale. On the other hand, the reason could also be that ChatGPT offered some causal explanations that led to answers mismatched with the labels, yet were convincing enough for the evaluation LLMs to consider them reasonable. 
These explanations might also make sense, considering the subjectivity involved in emotion-cause annotations.

We further conduct a case study as shown in Figure~\ref{fig:case_study}. In this case, the vanilla Vicuna performs the worst, with its reaction summary not aligning with the target sentence, leading to an unreasonable derivation of the stimulus content. Our multi-task model precisely identifies the direct factor, matching the final answer with the label and presenting a concise and straightforward reasoning chain. The ChatGPT seems to introduce some assumptions in the reaction part, but is considered to provide more detailed stimuli by GPT-4. 
Actually, in this example, the golden label corresponds to the most direct cause, while the additional two stimulus items provided by ChatGPT can be viewed as indirect background factors.
Therefore, if our goal is to extract more precise stimulus factors and acquire more direct reasoning chains, we may opt for the supervised model trained on our ECR-Chain set to avoid excessive irrelevant generated content. If we aim to discover more potential factors, we can directly instruct the LLM with our proposed ECR-Chains to fully leverage its rich knowledge and powerful reasoning capabilities, which may result in some insightful descriptions related to the emotion generation process.

\section{Conclusion}
In this paper, we explore emotion-cause reasoning with generative language models. We first introduce the Emotion-Cause Reasoning Chain (ECR-Chain), which can guide large language models (LLMs) to infer the emotion causes from emotional expressions with CoT prompting. Considering the limited reasoning abilities of smaller models, we utilize LLMs to automatically construct an ECR-Chain set through a four-stage processing workflow and perform supervised training based on the ECR-Chain set through multi-task learning. Experimental results demonstrate that our method can enhance the prediction of emotion-cause utterances while also enabling generative language models with the ability to perform explainable emotion-cause reasoning.

\appendix

\section*{Ethical Statement}
In this paper, we utilize ChatGPT to construct a reasoning chain dataset for supervised training. We notice that during the reasoning process, ChatGPT may generate hallucinations, such as conflating the identities of two speakers within a conversation. We propose revising strategies based on task labels and find that they can effectively remove some items that contain factual errors. However, there may still be some misleading content remaining in the set.

\section*{Acknowledgments}
This work was partially supported by the the National Natural Science Foundation of China (No. 62072462), National Key  R\&D  Program  of  China  (No.2020AAA0108600), and Beijing Natural Science Foundation (No. L233008).

\bibliographystyle{named}
\bibliography{refs}

\end{document}